\documentclass[conference, a4paper]{IEEEtran}
\IEEEoverridecommandlockouts

	\usepackage[turkish]{babel}
	\usepackage[utf8]{inputenc} 
	\usepackage[T1]{fontenc}

\usepackage{cite}

\ifCLASSINFOpdf
  \usepackage[pdftex]{graphicx}
\else
\fi
\usepackage[cmex10]{amsmath}
\usepackage{url}

\usepackage{multirow}
\usepackage{array}
\usepackage[lofdepth,lotdepth]{subfig}

\AtBeginDocument{%
  \renewcommand\tablename{TABLO}
}

\AtBeginDocument{%
  \renewcommand\abstractname{Abstract}
}

\begin{document}


%
\title{Büyük Dil Modelleri için TR-MMLU Benchmark’ı: Performans Değerlendirmesi, Zorluklar ve İyileştirme Fırsatları\\
TR-MMLU Benchmark for Large Language Models: Performance Evaluation, Challenges, and Opportunities for Improvement}

\author{
\IEEEauthorblockN{
M. Ali Bayram\IEEEauthorrefmark{1}, 
Ali Arda Fincan\IEEEauthorrefmark{2}, 
Ahmet Semih Gümüş\IEEEauthorrefmark{2}, 
Banu Diri\IEEEauthorrefmark{1},\\
Savaş Yıldırım\IEEEauthorrefmark{3},
Öner Aytaş\IEEEauthorrefmark{4},
}
\IEEEauthorblockA{\IEEEauthorrefmark{1}Yıldız Technical University, İstanbul, Turkey\\
Email: malibayram20@gmail.com, diri@yildiz.edu.tr}
\IEEEauthorblockA{\IEEEauthorrefmark{2}Yeditepe University, İstanbul, Turkey\\
Email: ardafincan@icloud.com, ahmetsemih3434@gmail.com}
\IEEEauthorblockA{\IEEEauthorrefmark{3}İstanbul Bilgi University, İstanbul, Turkey\\
Email: savasy@gmail.com}
\IEEEauthorblockA{\IEEEauthorrefmark{3}Işık University, İstanbul, Turkey\\
Email: oneraytas@gmail.com}
}

\maketitle

\IEEEpubid{\begin{minipage}{\textwidth}
  \vspace{50pt}
  \textbf{979-8-3315-6655-5/25/\$31.00 ©2025 IEEE}
\end{minipage}}

\begin{ozet}
Dil modelleri, insan dilini anlama ve üretme konularında önemli ilerlemeler kaydetmiş, birçok uygulamada dikkat çekici başarılar elde etmiştir. Ancak, özellikle Türkçe gibi kaynak açısından sınırlı dillere yönelik değerlendirme çalışmaları önemli bir zorluk oluşturmaktadır. Bu sorunu ele almak amacıyla, büyük dil modellerinin (LLM) Türkçe dilindeki dilsel ve kavramsal yeteneklerini değerlendirmek için kapsamlı bir değerlendirme çerçevesi olan Türkçe MMLU (TR-MMLU) benchmark'ını tanıttık. TR-MMLU, Türk eğitim sisteminden 62 bölümdeki 6.200 çoktan seçmeli soruyu içeren, özenle hazırlanmış bir veri setine dayanmaktadır. Bu benchmark, Türkçe doğal dil işleme (NLP) araştırmalarına standart bir çerçeve sunmakta ve büyük dil modellerinin Türkçe metinleri işleme yeteneklerini detaylı bir şekilde analiz etmeyi sağlamaktadır. Çalışmamızda, TR-MMLU üzerinde en güncel büyük dil modellerini değerlendirdik ve model tasarımında iyileştirme gerektiren alanları vurguladık. TR-MMLU, Türkçe NLP araştırmalarını ilerletmek ve gelecekteki yeniliklere ilham vermek için yeni bir standart oluşturmaktadır.
\end{ozet}

\begin{IEEEanahtar}
Büyük Dil Modelleri (LLM), Doğal Dil İşleme (NLP), Yapay Zeka, Türkçe NLP
\end{IEEEanahtar}

\begin{abstract}
Language models have made significant advancements in understanding and generating human language, achieving remarkable success in various applications. However, evaluating these models remains a challenge, particularly for resource-limited languages like Turkish. To address this issue, we introduce the Turkish MMLU (TR-MMLU) benchmark, a comprehensive evaluation framework designed to assess the linguistic and conceptual capabilities of large language models (LLMs) in Turkish. TR-MMLU is based on a meticulously curated dataset comprising 6,200 multiple-choice questions across 62 sections within the Turkish education system. This benchmark provides a standard framework for Turkish NLP research, enabling detailed analyses of LLMs' capabilities in processing Turkish text. In this study, we evaluated state-of-the-art LLMs on TR-MMLU, highlighting areas for improvement in model design. TR-MMLU sets a new standard for advancing Turkish NLP research and inspiring future innovations.
\end{abstract}




%
\IEEEpeerreviewmaketitle

\IEEEpubidadjcol

\section{G{\footnotesize İ}r{\footnotesize İ}ş}

Yapay zeka (AI) ve doğal dil işleme (NLP) alanındaki gelişmeler, özellikle GPT-4, BERT ve Llama gibi büyük dil modellerinin (LLM) ortaya çıkışıyla, hesaplamalı dilbilim alanında devrim yaratmıştır. Bu modeller, geniş veri kümeleriyle eğitilerek makine çevirisi, soru yanıtlama, içerik üretimi ve kod oluşturma gibi çeşitli uygulamalarda dikkat çekici yetenekler sergilemiştir. Ancak, bu modellerin gerçek yeteneklerini değerlendirmek, özellikle Türkçe gibi kaynak açısından sınırlı diller için hala büyük bir zorluk oluşturmaktadır \cite{hendrycks2021}.

Mevcut LLM değerlendirme kriterleri genellikle İngilizce gibi yaygın diller üzerine yoğunlaşmaktadır. Türkçe gibi eklemeli ve morfolojik açıdan zengin dillerin dilsel karmaşıklıkları ise çoğunlukla göz ardı edilmektedir. Türkçede bir fiil kökü, zamanlar, kişi ve kip gibi birçok bilgiyi içeren sayısız kelime formu oluşturabilir. Bu çeşitlilik, hem tokenizasyon hem de anlamsal çözümlemeyi karmaşık hale getirerek modellerin yüzeysel yapılar ve derin dilbilimsel yapılar arasında denge kurmasını gerektirir.

Türkçe NLP modellerini etkili bir şekilde değerlendirebilmek için bu dilin benzersiz özelliklerini dikkate almak gerekir. Türkiye’nin eğitim sisteminden türetilen sorular, bu tür değerlendirmeler için ideal bir temel sunmaktadır. Eğitim müfredatı, geniş kapsamlı konuları ve zengin dil içeriğiyle, dil modellerinin bilgi derinliğini ve dil yeteneklerini değerlendirmek için güçlü bir çerçeve sağlamaktadır. Bu sorular, yalnızca bilgi hatırlama değil, aynı zamanda kavramsal anlama, mantıksal akıl yürütme ve kültürel bağlamı test ederek Türkçe dil modellerini değerlendirmek için değerli bir kaynak oluşturmaktadır.

Büyük dil modellerinin değerlendirilmesi iki temel boyuta dayanır: \textit{talimat takibi} ve \textit{bilgi değerlendirme}. Talimat takibi, modellerin belirli komutları yerine getirme kabiliyetini ölçerken; bilgi değerlendirme, modellerin bilgi tabanının genişliğini ve derinliğini anlamayı hedefler. Türkçe için dilsel karmaşıklığın anlamsal nüansları gizleyebileceği göz önüne alındığında, bilgi değerlendirme, model performansını ölçmek için kritik bir metriktir.

Bu çalışma, Türkçe büyük dil modellerini değerlendirmede bilgi değerlendirmeyi öncelikli bir metrik olarak ele almaktadır. Çalışmada, TR-MMLU adlı yeni bir değerlendirme çerçevesi önerilmiş ve Türkiye'nin eğitim sistemindeki standart sınavlardan alınan sorular kullanılmıştır. Bu sınavlar, Türkçe NLP modellerini değerlendirmek için kültürel olarak uygun ve güçlü bir benchmark oluşturmaktadır.
\section{İlg{\footnotesize İ}l{\footnotesize İ} Çal{\footnotesize I}şmalar}

Büyük dil modellerinin hızlı gelişimi, bu modellerin farklı dilsel görevlerdeki performanslarını değerlendirmek amacıyla önemli çabaları beraberinde getirmiştir. MMLU, SuperGLUE ve SQuAD gibi benchmark'lar, özellikle İngilizce gibi yaygın konuşulan dillerde, anlama, bilgi derinliği ve üretkenlik yeteneklerini değerlendirmek için temel araçlar olmuştur. Ancak, Türkçe gibi kaynak açısından sınırlı diller, morfolojik zenginlikleri ve sentaktik karmaşıklıkları nedeniyle benzersiz zorluklar sunmaktadır. Bu doğrultuda, Türkçe için özel olarak geliştirilen çeşitli benchmark'lar, bu sınırlamaları ele almayı ve büyük dil modellerini anlamlı bir şekilde değerlendirmeyi amaçlamaktadır.

Türkçe NLP değerlendirme çalışmalarındaki en kapsamlı girişimlerden biri \textbf{TurkishMMLU} benchmark'ıdır. Bu benchmark, Türkçe LLM'lere yönelik yapılandırılmış bir değerlendirme çerçevesine olan ihtiyacı karşılamak için geliştirilmiştir. Veri seti, Türkiye'nin ulusal müfredatından türetilen soruları içermekte ve dilsel, kültürel ve kavramsal anlama yeteneklerini değerlendirmek için tasarlanmış çeşitli konuları kapsamaktadır. Her bir soru, gerçek dünya eğitim performansına dayalı bir zorluk derecesi ile derecelendirilerek model yeteneklerinin detaylı bir şekilde değerlendirilmesine olanak tanır. TurkishMMLU, Llama ve MT5 gibi açık kaynaklı çok dilli modellerin yanı sıra Türkçe uyarlamalı modeller üzerinde uygulanmış ve sıfır atış (zero-shot), az atış (few-shot) ve düşünce zinciri (chain-of-thought) akıl yürütme gibi farklı değerlendirme paradigmalarını desteklemektedir \cite{yuksel2024}.

Türkçe'ye özgü diğer benchmark'lar arasında \textbf{OpenLLMTurkishLeaderboard} ve güncellenmiş versiyonu olan \textbf{OpenLLMTurkishLeaderboard\_v0.2} bulunmaktadır. Bu benchmark'lar, MMLU, AI2\_ARC ve GSM8K gibi İngilizce benchmark'lardan çevrilen veri setlerine dayanmaktadır. Ancak, çeviri tabanlı benchmark'lar, Türkçe'nin dilsel ve kültürel nüanslarını yeterince yansıtamadığı için, modellerin yerel Türkçe görevlerdeki performanslarında tutarsızlıklara neden olabilir. Bununla birlikte, bu benchmark'lar üzerinde yapılan çalışmalar, Türkçe'ye özgü verilerle çok dilli modellerin ince ayarının (fine-tuning) etkili olduğunu göstermiştir \cite{openllm2024, dogan2024}.

\textbf{THQuAD} veri seti ise, tarihi Türkçe metinlere odaklanarak, BERT ve ELECTRA gibi modellerle okuma anlama görevlerini değerlendirmektedir. Bu veri seti, bağlama dayalı anlamayı ve tarihsel olarak köklü dilin yorumlanmasını içeren Türkçe NLP'ye özgü zorlukları vurgulamaktadır. Ancak, THQuAD'ın dar kapsamı, genel Türkçe NLP görevlerine uygulanabilirliğini sınırlandırmaktadır \cite{soygazi2021}.

Bu yapılandırılmış benchmark'ların yanı sıra, duygu analizi, soru yanıtlama ve dile özgü gömme vektörleri gibi Türkçe NLP görevlerine yönelik diğer çalışmalar da yürütülmüştür. Özellikle, duygu analizi araştırmaları, genellikle gayri resmi ve dilsel olarak çeşitli içerik barındıran Türkçe sosyal medya metinlerini işlemeye yönelik zorlukları ortaya koymuştur. Bu çalışmalar, Türkçe LLM'ler ile İngilizce modeller arasındaki performans farkını kapatmak için kültürel olarak uyumlu, Türkçe'ye özgü veri setlerinin kritik önemini vurgulamaktadır.

Mevcut benchmark'larla karşılaştırıldığında, Türkçe MMLU (TR-MMLU) benchmark'ı aşağıdaki temel avantajları sunmaktadır:
\begin{itemize}
    \item TR-MMLU, Türkçe için özgün olarak tasarlanmıştır ve çok dilli benchmark'larda sıklıkla görülen çeviri hataları ve kültürel uyumsuzlukları ortadan kaldırır.
    \item Veri seti, sağlık, hukuk, tarih ve doğa bilimleri gibi geniş bir konu yelpazesini kapsayarak, Türkçe'nin dilsel ve kavramsal çeşitliliğini yansıtan bütünsel bir değerlendirme çerçevesi sağlar.
    \item TR-MMLU, hem talimat takibi yeteneklerini hem de bilgi anlama yeteneklerini değerlendirerek, Türkçe LLM'lerin kapsamlı bir şekilde analiz edilmesini mümkün kılar.
\end{itemize}

Sonuç olarak, TurkishMMLU, OpenLLMTurkishLeaderboard ve THQuAD gibi mevcut benchmark'lar, Türkçe NLP'nin temellerini atmıştır, ancak dilsel derinlik ve değerlendirme kapsamı açısından önemli boşluklar bırakmaktadır. TR-MMLU, bu boşlukları doldurarak, Türkçe'ye özgü, şeffaf ve kültürel olarak uyumlu bir çerçeve sunmaktadır. TR-MMLU, model değerlendirme ve geliştirme alanında ilerlemeler sağlayarak, Türkçe NLP'de gelecekteki araştırma ve yenilikler için yeni bir standart belirlemektedir.
\section{G{\footnotesize Ö}rev Tan{\footnotesize I}m{\footnotesize I} ve Y{\footnotesize Ö}ntem}

Bu çalışmanın temel amacı, Türkçe dilinde Büyük Dil Modellerinin performansını değerlendirmek için kapsamlı bir benchmark oluşturmaktır. Türkçe MMLU benchmark'ı (TR-MMLU), 62 kategoriye yayılan 6.200 çoktan seçmeli sorudan oluşan yüksek kaliteli bir veri seti kullanılarak geliştirilmiştir. Bu kategoriler, hukuk, sağlık, tarih ve sanat gibi çeşitli alanları kapsamakta ve sorular, Türk eğitim sistemi ile diğer uzmanlık alanlarından özenle türetilmiştir. TR-MMLU, Türkçe doğal dil işleme araştırmalarındaki kritik boşlukları ele alarak dilsel ve kavramsal anlayışı değerlendirmek için standart ve şeffaf bir çerçeve sağlamayı hedeflemektedir.

TR-MMLU'nun üç temel amacı bulunmaktadır:
\begin{itemize}
    \item Farklı konular üzerindeki anlama düzeyini nesnel olarak ölçerek LLM'lerin güçlü ve zayıf yönlerine dair içgörüler sunmak.
    \item Tüm soruları, cevapları ve değerlendirme betiklerini kamuya açık hale getirerek şeffaflık ve tekrarlanabilirlik sağlamak.
    \item Model performansındaki boşlukları tespit ederek, doğru ve sağlam Türkçe dil modellerinin geliştirilmesine rehberlik etmek.
\end{itemize}

TR-MMLU’nun önemi, Türkçe NLP’nin karşılaştığı özel zorlukları ele alma yeteneğinde yatmaktadır. Çevirilere dayalı mevcut çok dilli benchmark’ların aksine, TR-MMLU Türkçe alanında uzmanlar tarafından hazırlanmıştır ve kültürel ve dilsel uyumluluğu sağlayarak çeviri kaynaklı hataları ortadan kaldırır. Ayrıca, veri seti ön eğitim verileriyle örtüşmeyecek şekilde özenle düzenlenmiştir, bu da modellerin performansını önceden edinilmiş bilgiden bağımsız olarak değerlendirmeyi mümkün kılar.

TR-MMLU değerlendirme yöntemi, tutarlılığı ve doğruluğu sağlamak için dikkatle tasarlanmıştır. Toplamda 39 LLM değerlendirilmiş, açık kaynaklı modeller (örneğin, Llama, Gemma) ve kapalı kaynaklı modeller (örneğin, GPT-4, Claude) dahil edilmiştir. Değerlendirme süreci, karşılaştırılabilir sonuçlar sağlamak için kontrollü donanım ortamlarında gerçekleştirilmiş ve doğruluk, başarı oranı ve işlem süresi gibi metrikler kullanılarak model performansı analiz edilmiştir. Cevap formatlarındaki farklılıkları ele almak için parafraz algılama modelleri kullanılmış ve yanıtlar anlam açısından tutarlı hale getirilmiştir.

Model yanıtlarının doğruluğunu artırmak için "prompt engineering" önemli bir rol oynamıştır. Türkçe'nin dilsel özelliklerine ve değerlendirilen modellerin yeteneklerine uygun çeşitli prompt yapıları test edilmiştir. Sonuçlar, farklı prompt koşullarında verilen doğru yanıt sayısına göre analiz edilerek, yönlendirme stratejilerinin etkinliği hakkında içgörüler sunmuştur.

Hugging Face platformunda kamuya açık bir liderlik tablosu oluşturulmuş ve modeller, doğruluk ve performans metriklerine göre sıralanmıştır. Bu liderlik tablosu, her model hakkında mimari, parametre boyutu ve kuantizasyon seviyesi gibi ayrıntılı bilgiler sunarak, araştırma topluluğunun model performansını karşılaştırması ve en iyi uygulamaları belirlemesi için güvenilir bir referans sağlamaktadır.

TR-MMLU, Türkçe NLP alanında, Türkçenin benzersiz özelliklerine uygun bir benchmark sunarak önemli bir ilerleme kaydetmektedir. Şeffaflık, tekrarlanabilirlik ve iş birliğini teşvik ederek, TR-MMLU yalnızca Türkçe dil modellerini değerlendirmek için yüksek bir standart belirlemekle kalmamakta, aynı zamanda kaynak açısından sınırlı diller için NLP’de gelecekteki yeniliklerin temellerini atmaktadır.
\section{Deneyler ve Sonu{\footnotesize Ç}lar}

Deneyler, TR-MMLU veri setindeki 6.200 soruyu 39 büyük dil modeli üzerinde değerlendiren bir Python betiği kullanılarak Ollama platformunda gerçekleştirilmiştir. Tekrarlanabilirliği sağlamak amacıyla tüm testler sabit bir 42 rastgele tohum değeri ile yürütülmüştür. Yanıt doğruluğunu artırmak için Türkçe'ye özgü çeşitli yönlendirme (prompt) yapıları kullanılmıştır. Aşağıda değerlendirme sırasında kullanılan örnek yönlendirmeler verilmiştir:

\textbf{Yönlendirme Örnekleri:}
\begin{itemize}
    \item \textbf{Yönlendirme 1:} \textit{Sana soru ve seçenekleri veriyorum. Sadece hangi seçeneğin sorunun doğru cevabı olduğunu yaz.}  
    Sonuçlar: \texttt{gemma2:9b = 63 doğru, llama3.1 = 47 doğru}
    
    \item \textbf{Yönlendirme 2:} \textit{Sana çoktan seçmeli soru ve seçeneklerini veriyorum. Sorunun doğru seçeneğini bul ve sadece doğru seçeneğin hangi şıkka ait olduğunu söyle.}  
    Sonuçlar: \texttt{gemma2:9b = 60 doğru, llama3.1 = 33 doğru}
    
    \item \textbf{Yönlendirme 3:} \textit{Sana soru ve seçenekleri veriyorum, sorunun cevabının hangi seçenek olduğunu bul ve sadece doğru seçeneğin hangi şıkka ait olduğunu söyle.}  
    Sonuçlar: \texttt{gemma2:9b = 58 doğru, llama3.1 = 37 doğru}
    
    \item \textbf{Yönlendirme 4:} \textit{Sana vereceğim çoktan seçmeli sorunun sadece doğru şıkkının harfini söyle.}  
    Sonuçlar: \texttt{gemma2:9b = 55 doğru, llama3.1 = 36 doğru}
    
\end{itemize}

Değerlendirme sonuçları Hugging Face platformunda üç ayrı veri seti olarak yayımlanmıştır:
\begin{enumerate}
    \item \textbf{Yapay Zeka Türkçe MMLU Liderlik Tablosu:} Bu veri seti, doğruluk, parametre boyutu, kuantizasyon seviyesi ve işlem süresi gibi metriklere göre modellerin sıralamasını içermektedir. Tablo~\ref{tab:leaderboard}, seçili modellerin performansını özetlemektedir.

    \begin{table}[h]
  \centering
  \caption{\small\textsc{Yapay Zeka Türkçe MMLU Liderlik Tablosu}}
  \label{tab:leaderboard}
  \begin{tabular}{|l|c|c|c|c|c|c|}
    \hline
    \textbf{Model} & \textbf{Aile} & \textbf{Parametre} & \textbf{Kuant.} & \textbf{Doğru} & \textbf{Doğruluk} & \textbf{Süre} \\
    & & \textbf{Boyutu} & & \textbf{Cevaplar} & (\%) & (s) \\
    \hline
    gpt-4o & GPT & Bilinmiyor & Yok & 5260 & 84.84 & 5021 \\
    \hline
    claude-3.5 & Sonnet & Bilinmiyor & Yok & 5233 & 84.40 & 7379 \\
    \hline
    llama3.3:latest & llama & 70.6B & Q4\_K\_M & 4924 & 79.42 & 13355 \\
    \hline
    gemini-1.5-pro & Gemini & Bilinmiyor & Yok & 4758 & 76.74 & 4985 \\
    \hline
    gemma2:27b & gemma2 & 27.2B & Q4\_0 & 4470 & 72.10 & 5506 \\
    \hline
  \end{tabular}
\end{table}
    \item \textbf{Yapay Zeka Türkçe MMLU Bölüm Sonuçları:} Bu veri seti, 62 kategori genelindeki detaylı performans analizlerini içermekte ve modellerin güçlü ve zayıf yönlerini vurgulamaktadır. Tablo~\ref{tab:section_results}, Tıpta Uzmanlık Sınavı (TUS) ve Kamu Personeli Seçme Sınavı (KPSS) gibi temel alanlarda seçili modellerin performansını göstermektedir.

\begin{table}[h]
  \centering
  \caption{\small\textsc{Yapay Zeka Türkçe MMLU Bölüm Sonuçları (Seçili Kategoriler)}}
  \label{tab:section_results}
  \begin{tabular}{|l|c|c|c|c|c|}
    \hline
    \textbf{Model} & \textbf{Genel} & \textbf{TUS} & \textbf{KPSS} & \textbf{Ehliyet} & \textbf{AÖF} \\
    & \textbf{Ort. (\%)} & (\%) & (\%) & (\%) & \textbf{Ort. (\%)} \\
    \hline
    gpt-4o & 84.84 & 91 & 74.5 & 97 & 84.55 \\
    \hline
    claude-3.5 & 84.40 & 88 & 71.5 & 96 & 84.65 \\
    \hline
    llama3.3:latest & 79.42 & 85 & 66.5 & 92 & 79.58 \\
    \hline
    gemma2:27b & 72.10 & 77 & 60 & 90 & 72.57 \\
    \hline
    aya-expanse:32b & 70.66 & 69 & 55.5 & 84 & 70.96 \\
    \hline
  \end{tabular}
\end{table}
    
    \item \textbf{Yapay Zeka Türkçe MMLU Model Yanıtları:} Bu veri seti, tüm modellerin detaylı cevaplarını içererek hata analizi ve model davranışlarının daha derinlemesine anlaşılmasını sağlamaktadır.
\end{enumerate}

Yanıt formatındaki farklılıkları ele almak için “paraphrase-multilingual-mpnet-base-v2” modeli kullanılmıştır. Bu model, üretilen yanıtlar ile doğru cevaplar arasındaki anlamsal benzerlik skorlarını hesaplamış ve en yüksek puanlı seçeneği doğru kabul etmiştir.

Sonuçlar, Türkçe morfolojisine uygun güçlü tokenizasyon stratejilerine sahip modellerin diğerlerinden daha iyi performans gösterdiğini ortaya koymuştur. İnce ayar yapılan (fine-tuned) modeller önemli performans artışları sağlasa da, \textit{catastrophic forgetting} gibi sorunlar gözlemlenmiştir. Bu bulgular, yerelleştirilmiş veri setlerinin ve optimize edilmiş eğitim stratejilerinin önemini vurgulamaktadır.

TR-MMLU'yu çoktan seçmeli soruların ötesine genişleterek açık uçlu görevler, duygu analizi ve adlandırılmış varlık tanıma gibi alanlara dahil etmek, bu benchmark'ın kullanımını daha da artıracaktır. Ayrıca, çeşitli ve yüksek kaliteli Türkçe veri setlerinin geliştirilmesi, Türkçe NLP'nin ilerlemesi için kritik bir öncelik olmaya devam etmektedir.
\section{Sonu{\footnotesize Ç}}

TR-MMLU benchmark'ı kullanılarak gerçekleştirilen Türkçe büyük dil modellerinin değerlendirilmesi, Türkçe doğal dil işleme alanını ilerletmek için çeşitli zorluklar ve fırsatlar ortaya koymuştur. En önemli zorluklardan biri, Türkçe’nin eklemeli (agglutinative) yapısı ve karmaşık morfolojisi nedeniyle ortaya çıkan tokenizasyon problemidir. Mevcut modellerin birçoğu, Türkçe tokenleri etkili bir şekilde işleyememekte ve bu durum doğruluk oranlarının düşmesine neden olmaktadır. Gelecekteki çalışmalar, Türkçe’ye özgü dilsel karmaşıklıkları ele alan ve daha doğru model temsilleri sağlayan gelişmiş tokenizasyon tekniklerine odaklanmalıdır.

Türkçe'ye özel görevler için kullanılan ince ayar (fine-tuning) stratejileri, iyileştirme alanı sunan bir diğer konudur. Türkçe veri setleri üzerinde yapılan ince ayar işlemleri performansı artırsa da, genellikle \textit{catastrophic forgetting} olarak bilinen, daha önce öğrenilen bilginin kaybolması gibi zorluklarla karşılaşılmaktadır. Bu sorunu azaltmak için, temel bilgiyi korurken Türkçe’ye özel görevler için modelleri optimize eden daha dayanıklı ince ayar yöntemleri araştırılmalıdır.

Yüksek kaliteli Türkçe veri setlerinin sınırlı olması, kapsamlı model değerlendirme ve eğitimine engel teşkil eden bir diğer sorundur. Çeşitli ve etik olarak elde edilmiş Türkçe veri setlerinin havuzunu genişletmek, yalnızca ince ayar süreçlerini desteklemekle kalmayacak, aynı zamanda çeşitli NLP uygulamalarında genellenebilirliği artıracaktır. Ayrıca, TR-MMLU benchmark'ını açık uçlu görevler, duygu analizi, adlandırılmış varlık tanıma ve bağlamsal kelime gömme gibi alanları da içerecek şekilde genişletmek, Türkçe dil modellerinin daha bütüncül bir değerlendirmesini sağlayacaktır.

Farklı kategorilerde gözlemlenen model performansı farklılıkları, bu varyasyonların temel nedenlerini daha derinlemesine araştırma gereğini ortaya koymaktadır. Bu tür araştırmalar, Türkçe LLM’lerin güçlü ve zayıf yönlerini netleştirecek ve hedefe yönelik iyileştirmeler için yol gösterecektir. Bu sayede modeller, Türkçe’nin dilsel nüanslarına daha iyi uyum sağlayacak ve farklı uygulamalarda çok yönlülüklerini artıracaktır.

TR-MMLU, Türkçe NLP için ileriye doğru atılmış önemli bir adımdır ve Türkçe dili için özel olarak tasarlanmış sağlam ve şeffaf bir değerlendirme çerçevesi sunmaktadır. Türk eğitim sistemi ve uzmanlık alanlarından türetilen 6.200 çoktan seçmeli soru ile bu benchmark, araştırmacıların ve geliştiricilerin model performansını nesnel olarak ölçmesi için değerli bir kaynak işlevi görmektedir.

Bu çalışmanın bulguları, tokenizasyon ve dile özgü ince ayar işlemlerinin kritik önemini vurgulamaktadır. Türkçe morfolojisine uygun gelişmiş tokenizasyon tekniklerine sahip modellerin daha yüksek doğruluk oranlarına ulaştığı gözlemlenmiştir. Ayrıca, Türkçe’ye özgü verilerle yapılan ince ayar işlemleri önemli performans artışları sağlamış, ancak temel bilginin korunması konusunda bazı sınırlamalar da ortaya çıkmıştır. Bu sonuçlar, Türkçe NLP’nin benzersiz zorluklarını ele almak için model tasarımı ve eğitiminde özel yaklaşımların gerekliliğini ortaya koymaktadır.

Şeffaflık ve tekrarlanabilirliğe vurgu yapan güvenilir bir benchmark olarak TR-MMLU, araştırma topluluğu içinde iş birliği ve yenilikleri teşvik etmektedir. Gelecekte, benchmark çoktan seçmeli soruların ötesine geçerek daha çeşitli değerlendirme görevlerini kapsayacak şekilde genişlemeyi hedeflemektedir. Bu genişleme, yalnızca TR-MMLU’nun kullanımını artırmakla kalmayacak, aynı zamanda kaynak açısından sınırlı diller için dil modeli değerlendirme standartlarını da yeniden tanımlayacaktır.

Sonuç olarak, TR-MMLU, Türkçe NLP için temel bir ilerleme sunmakta ve dil modellerini değerlendirmek için standart bir çerçeve sağlamaktadır. Tokenizasyon, ince ayar ve veri seti erişilebilirliği gibi temel zorlukları ele alarak ve değerlendirme metriklerini genişleterek, araştırmacılar ve geliştiriciler Türkçe dil işleme teknolojilerini daha da geliştirmeye teşvik edilmektedir. Bu gelişmeler, alanda yeni standartlar belirlemeye ve daha sağlam ve çok yönlü Türkçe LLM’lerin geliştirilmesine katkıda bulunacaktır.

\end{document}